# GAC-Net：Geometric and attention-based Network for Depth Completion

Kuang Zhu[1] ·· Min Sun[2] · Xingli Gan[*]

212308802004@zust.edu.cn

Abstract:
Depth completion is a key task in autonomous driving, aiming to complete sparse LiDAR depth measurements into high-quality dense depth maps through image guidance. However, existing methods usually treat depth maps as an additional channel of color images, or directly perform convolution on sparse data, failing to fully exploit the 3D geometric information in depth maps, especially with limited performance in complex boundaries and sparse areas. To address these issues, this paper proposes a depth completion network combining channel attention mechanism and 3D global feature perception (CGA-Net). The main innovations include: 1) Utilizing PointNet++ to extract global 3D geometric features from sparse depth maps, enhancing the scene perception ability of low-line LiDAR data; 2) Designing a channel-attention-based multimodal feature fusion module to efficiently integrate sparse depth, RGB images, and 3D geometric features; 3) Combining residual learning with CSPN++ to optimize the depth refinement stage, further improving the completion quality in edge areas and complex scenes. Experiments on the KITTI depth completion dataset show that CGA-Net can significantly improve the prediction accuracy of dense depth maps, achieving a new state-of-the-art (SOTA), and demonstrating strong robustness to sparse and complex scenes.

## 1.INTRODUCTION

Depth completion technology [1] plays a key role in computer vision applications such as autonomous driving [2], scene understanding [3], 3D reconstruction [4], object detection [5], and pose estimation [6]. For autonomous driving, depth completion can process relatively low-cost LiDARs (such as 64-line LiDARs) through algorithms to approximate the measurement accuracy of high-line LiDARs, thereby effectively improving the accuracy and stability of perception of pedestrians, vehicles, and other objects. Specifically, depth completion [1] aims to recover pixel-level dense depth maps from sparse and noisy depth measurement data, which can be assisted by high-resolution RGB images or completed independently.

In outdoor scenes, LiDAR can only provide relatively sparse but accurate depth values. The mainstream approach usually enriches texture and color features with the help of synchronously captured RGB images. However, a large number of studies [7][8][9] have shown that images themselves have inherent depth ambiguity, which can easily produce unstructured depth features. At the same time, how to fully exploit the scene-level structural information contained in sparse depth measurement data is still one of the difficulties that depth completion needs to overcome.

One-stage multi-modal fusion methods based on deep learning [10][11] can recover dense depth to a certain extent, but often lead to excessive smoothing of edges and loss of details. To solve this problem, researchers have added a propagation-based post-processing stage after the multi-modal fusion network, forming a two-stage network structure [12][13][14], aiming to alleviate excessive smoothing. However, such two-stage methods still have deficiencies in distinguishing between valid and invalid measurement points [15] and processing irregularly sampled sparse depth points [16], leading to limited overall performance. Recently, some scholars [17] have added a pre-processing step before the two-stage method, first obtaining an initial dense depth map, and then handing it over to the multi-modal fusion network for more accurate completion, achieving better results. But since the three-dimensional structural information in the depth channel has not been fully utilized, the performance of these methods in complex boundaries and sparse areas is still not ideal.

In response to this challenge, some works have tried to guide 2D convolution with depth values [18][19], but simply treating the depth map as an additional 2D channel cannot fully exploit its inherent structure [17]. Therefore, we propose a method of using the 3D geometric prior of the depth map to improve the completion accuracy: First, the PointNet++ [20] encoder is used to extract global features from the sparse point cloud. Only one feature calculation is needed at the initial stage of the network to capture the global geometric information at the scene level, which significantly improves the completion quality of sparse areas and large holes, while reducing the repeated calculation cost of the multi-scale stage. Subsequently, a feature fusion module based on channel attention mechanism [21] is introduced in multi-modal fusion. The U-Net [22] is used to perform early fusion of RGB images and initial dense depth maps, and then explicitly combined with 3D global features. It can adaptively adjust the feature weights and improve the completion accuracy in complex edge areas. Finally, in the depth refinement stage, residual learning is used to guide the fine propagation of CSPN++ [12], combined with the valid measurement points of the sparse depth map, to further reduce error accumulation and improve edges and local details.

Therefore, compared with relying solely on traditional multi-modal fusion or two-stage solutions, this work realizes higher perception accuracy and robustness in autonomous driving scenarios by mining the geometric priors contained in high-line LiDAR from the sparse data of low-line LiDAR. Our main contributions are as follows:

Global 3D feature extraction: Introducing PointNet++ [20] to extract scene-level geometric priors from sparse point clouds greatly improves the depth completion effect of sparse areas and large holes.

Multi-modal fusion based on channel attention: Designing a channel attention mechanism [21] to adaptively fuse image features with 3D global features significantly improves the completion accuracy in complex scenes.

Residual learning and CSPN++ depth refinement: Combining residual learning and CSPN++ [12] propagation can greatly reduce error accumulation and optimize edges and local details while maintaining high-precision completion.

## 2.RELATED WORK

In this section, we briefly review two related topics, including depth completion、2D-3D Joint Depth Completion.

1.Depth Completion

Early depth completion methods largely relied on traditional image processing techniques [23][24], such as bilateral filtering and interpolation, to fill holes and denoise relatively dense depth maps from structured light sensors based on structures. However, these methods performed poorly when dealing with sparse depth maps generated by devices such as LiDAR, especially in outdoor scenes. With the development of deep learning technologies, deep learning-based depth completion methods have emerged. These methods mainly use semantic information provided by RGB images to guide the completion of sparse depth maps. Methods like Enet [25] extract image features and depth features through separate encoders, and then fuse these multi-modal features to predict dense depth maps. CSPN [26] and its improved version CSPN++ [12] use convolutional spatial propagation networks to utilize the sparse but accurate data revisited from the original depth maps to enhance boundary details and global consistency. Methods such as BPNet [17] introduce a preprocessing stage to generate initial dense depth maps from the original sparse depth maps containing irregular data points, and then perform multi-modal fusion, improving the effect of depth completion. In recent years, researchers have gradually focused on introducing 3D representations to complete depth completion tasks. Methods like GraphCSPN [27] and KBNet [28] convert sparse depth data into point clouds and then use point cloud-based feature extraction modules to capture geometric information in the scene. However, due to the sparsity of point cloud data, these methods still face great challenges in 3D feature learning. Therefore, existing depth completion methods have not fully exploited the processing and utilization of 3D features. Our network extracts global geometric features from sparse depth maps through PointNet++ [20], mining more information hidden in the depth channel.

2.2D-3D Joint Depth Completion

Most existing mainstream depth completion methods are based on image guidance. In order to fully utilize the information in sparse depth maps, in recent years, researchers have begun to combine 3D representations on the basis of image features to supplement and enhance the representation of image-guided methods. DLiDAR and DepthNormal first introduced surface normals to improve the effect of depth completion. Considering the ability of graph neural networks to process data such as point clouds, ACMNet uses attention-based graph propagation for multi-modal fusion. Furthermore, GraphCSPN [27] combines graph neural networks and convolutional spatial propagation networks for depth completion. In addition, FuseNet [32] and PointDC [33] model 3D geometry from LiDAR point clouds and then incorporate it into the network for processing. We use a feature fusion module based on channel attention mechanism to supplement global 3D features after simply fusing image features with depth map features, more fully exploring the fusion potential of 2D and 3D features.

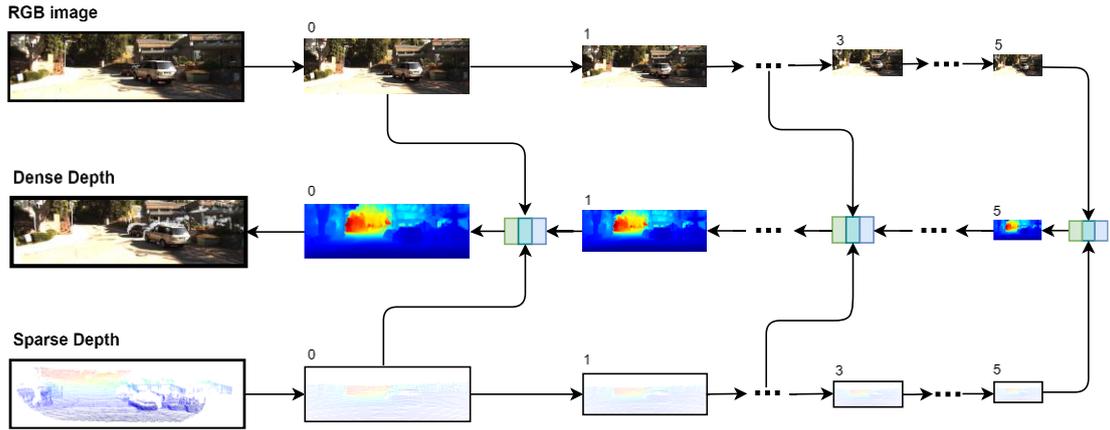

Figure 1 Multiscale Network Structure: Estimating dense depth through a six-scale network, from low resolution (scale 5) to high resolution (scale 0). At each scale, dense depth maps are generated from sparse depth maps and image features through the three-stage processing described in Figure 2.

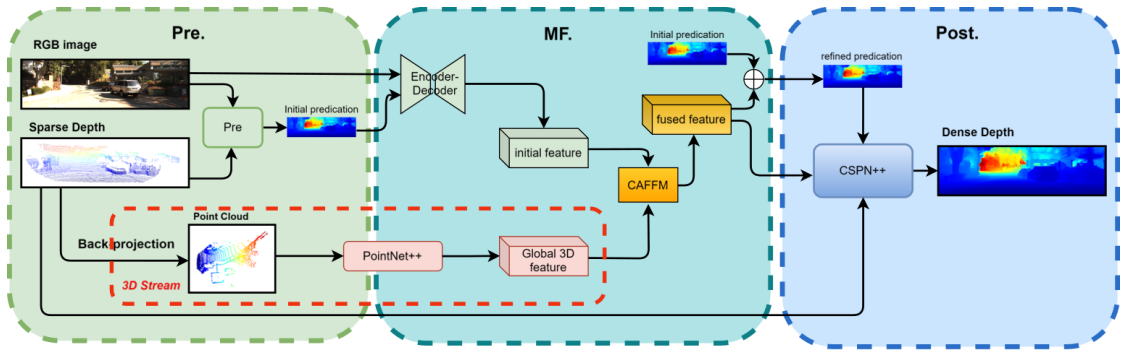

Figure 2 Overview of the specific three-stage network. At each scale, it consists of three consecutive stages, namely: the Preprocessing stage (Pre.), the Multimodal Fusion stage (MF.), and the Depth Refinement stage (Post.). Based on the two-stage network, we have added a preprocessing module to provide an initial dense depth map for multimodal fusion. Moreover, our network has a 3D stream branch, which extracts 3D features of the point cloud through PointNet++ to obtain a global 3D feature vector, and then further fuses features using the Channel Attention Feature Fusion Module (CAFFM) to enhance the effect of multimodal fusion.

### 3. OUR APPROACH

1.overview

Given a pair of input data: a color image $I$ and a sparse depth map $S$ our goal is to generate a dense depth map $D$ through a depth completion network. The color image $I$ provides rich texture details and color information, while the sparse depth map $S$ contains relatively accurate but irregular depth information. The objective of depth completion is to predict a complete dense depth map by fusing these complementary pieces of information.

We propose a six-scale, three-stage multimodal fusion depth completion network. As shown in Figure 1, the entire network architecture consists of six scales (from the lowest resolution scale 0 to the highest resolution scale 5). As illustrated in Figure 2, each scale is composed of three stages (preprocessing stage, multimodal fusion stage, and depth refinement stage). In the preprocessing stage, inspired by BPNet, we introduce a bilateral propagation module to generate an initial dense depth map, preparing for the multimodal fusion stage. In the multimodal fusion stage, a simple U-Net is used to fuse image features and depth features of the initial depth map in an early fusion manner to obtain preliminary fused features. Then, a feature fusion module based on channel attention mechanism adaptively fuses the global 3D features extracted by PointNet++ and the preliminary fused features, thereby obtaining more representative fused features in the multimodal fusion stage. In the depth refinement stage, we use CSPN++ to process the fused features generated by the multimodal fusion stage, correct the dense depth map, and the original sparse depth map, updating the dense depth map with sparse depth through a convolutional propagation module in an iterative manner.

We introduce the details of the 3D feature extraction module, preprocessing stage, multimodal fusion stage, depth refinement stage, six-scale network architecture, and loss function in Sections 3.2 to 3.7.

2. 3D Feature Extraction Module

Since the scene-level structural prior of sparse depth is crucial for improving the accuracy of depth completion, we propose a 3D feature extraction module to extract the geometric information hidden in the depth map. We use intrinsic parameters to back-project the valid data in the sparse depth map into 3D space to obtain the corresponding point cloud representation: $P_S$. Since PointNet++ can recursively aggregate local features and increase contextual scales, thereby efficiently capturing the geometric features of the point cloud, we choose to modify the point-based PointNet++ network, using only its encoder part as a feature extractor to extract the global 3D features of the point cloud $P_S$. To avoid repeatedly constructing point clouds and extracting features, we save the extracted global 3D features as a global vector, which is reused in each scale, significantly improving computational efficiency.

3. Preprocessing Stage

In the field of depth completion, 2-Stage methods used to be mainstream. These methods typically include a multimodal fusion stage and a post-processing stage. However, these methods face significant challenges when dealing with sparse depth maps, especially in outdoor scenes where LiDAR often generates irregular and unstable sparse data. Relying solely on methods such as sparse convolution may not be sufficient to address these issues effectively. Therefore, we have added a preprocessing stage, inspired by BPNet. In the preprocessing stage, a bilateral propagation module is used to initially densify the sparse depth map, providing more reliable input for the multimodal fusion stage and reducing the uncertainty brought by sparsity. Moreover, the bilateral propagation module dynamically generates propagation weights by combining information about image content and local spatial distances, making the densification process more adaptive to the non-uniform distribution of sparse data.

4. Multimodal Fusion Stage

In the multimodal fusion stage of the depth completion network, we design a Channel Attention Feature Fusion Module (CAFFM) based on channel attention mechanism, in addition to the simple fusion of image features and depth features, to fully utilize the rich texture information of the image, the depth features of the depth map, and the 3D global geometric features extracted from the point cloud. The specific design is as follows:

1.Early Fusion

In the multimodal fusion stage of each scale, we first adopt an early fusion approach, concatenating the image features and the depth features from the initial dense depth map (generated by the preprocessing stage) along the channel dimension. Then, we use a U-Net to process the concatenated feature map, leveraging its encoder-decoder structure to achieve feature fusion. This method generates a preliminary fused feature map that combines the rich texture information of the image and the geometric information from the depth data, laying the foundation for subsequent feature enhancement.

2.Feature Fusion Based on Channel Attention

Building on the preliminary fusion, we further design a Channel Attention Feature Fusion Module (CAFFM) to fuse the 3D global features extracted from the point cloud with the preliminary fused features output by the U-Net.

(1)Channel Attention Mechanism: Through the Squeeze-and-Excitation structure (SE module), it dynamically learns the weights of each channel, explicitly introducing 3D global features to enhance the expressiveness of the features.

For two input features $F_{U-Net}$ and $F_{3D}$, global pooling operations ($GlobalPool$) are performed respectively to generate global features. Then, a fully connected layer ($FC$) is used to generate the attention weights $Attention_{U-Net}$ and $A_{3D}$ for each feature, as shown in the following formulas:

$$Attention_{U-Net} = \sigma(FC(GlobalPool(F_{U-Net})))$$

$$Attention_{3D} = \sigma(FC(GlobalPool(F_{3D})))$$

$F_{U-Net}$ is the $U-Net$ initially fused feature output, $F_{3D}$ is the 3D global feature extracted by PointNet++, $Attention_{U-Net}$ is the attention weight for the initial fusion feature, and $GlobalPool$ is the attention weight vector for the 3D global feature. $FC$ is the fully connected layer, and $\sigma$ is the activation function used to compress the weight values into the range $[0,1]$.

(2) Weighted Fusion: The generated attention weight vectors are used for weighting and fusion.

$$F_{fused} = (A_{U-Net} \cdot F_{U-Net}) + (A_{3D} \cdot F_{3D})$$

(3) Fused Feature Representation: The final fused feature $F_{fused}$ integrates the texture information from the image, the depth information from the initial dense depth map, and the 3D global geometric prior information, forming a feature map with stronger expressiveness. The final fused feature is input into the subsequent depth refinement module, where CSPN++ is used to further enhance the completion accuracy of the dense depth map in sparse areas and complex boundaries.

## 5. Post-processing Stage

The post-processing stage, also known as the depth refinement stage, is based on the classic and mature convolutional spatial propagation module. It optimizes the corrected depth map generated in the multimodal fusion stage to further refine the depth information. The following are the specific steps and formulas for the depth refinement stage:

(1) Initial Depth Map Correction

First, use the fused features $F_{fused}$ generated in the multimodal fusion stage to create residual features, which are then used to correct the initial dense depth map $\hat{D}_{pre}$ generated in the preprocessing stage:

$$\hat{D}_0 = \hat{D}_{pre} + \Delta D$$

(2) Depth Propagation Update

The corrected initial depth map $D_0$ is input into the improved module for multi-step depth propagation optimization. The update formula for the propagation process during the $t$ iteration is:

$$\hat{D}_{i,t+1} = k_i \hat{D}_{i,t} + \sum_{j \in N(i)} k_{i,j} \hat{D}_{j,t}$$

Here:

$$k_{ij} = \frac{\hat{k}_{ij}}{\sum_{j \in N(i)} |\hat{k}_{ij}|}$$

$$k_i = 1 - \sum_{j \in N(i)} k_{ij}$$

$N(i)$ represents the local neighborhood of pixel $i$, $\hat{k}_{ij}$ is the content-related affinity weight generated through the fused feature $F_{fuesd}$.

(3) Sparse Depth Constraint

To preserve the depth information of valid points in the sparse depth map, we constrain the propagation results in each iteration, with the formula:

$$\hat{D}_{i,t+1} = (1-\gamma_i)\hat{D}_{i,t+1} + \gamma_i S_i$$

Where $\gamma_i = \sigma(Conv(F_{fused}))$ represents the confidence weight, $S_i$ is the valid depth value in the sparse depth map, and $\cdot \sigma$ denotes the multiplication operation.

## 4. Final Deep Fusion

After multiple propagation steps, the depth refinement results are fused through a multi-kernel

multi-step propagation mechanism to ultimately generate the optimized depth map:

$$D_i = \sum_{k \in K} \sum_{t \in T} \tau_{k,t} \hat{D}_{i,k,t}$$

Here, $K$ represents a set of different kernel sizes $\{3,5,7\}$, $T$ represents a set of different iteration steps, and $\tau_{k,t}$ are the weight parameters generated based on the fused feature $F_{fused}$.

6. Six-Scale Network Architecture

Inspired by the multi-scale cascaded hourglass network [MSG-CHN] [34], our entire network architecture adopts a six-scale multi-scale scheme, as shown in Figure 1. Each scale consists of a preprocessing stage, a multimodal fusion stage, and a post-processing stage. We accept feature maps and depth maps of specific resolutions at each scale and generate prediction results of the same resolution. Scales with lower resolutions can better capture coarse structures and grasp the overall prediction level, while scales with higher resolutions use the results of lower resolutions as priors to produce finer-grained predictions. In the entire framework, the resolutions of each scale are $1/32$, $1/16$, $1/8$, $1/4$, $1/2$, and $1/1$ of the original resolution, respectively.

7.Loss Function

Given that our network is a multi-scale architecture with six scales, we opt for a multi-scale loss based on L2 loss to supervise the depth maps at each scale. Since the ground truth contains invalid pixels, only those pixels with valid depth values are considered. The loss function is defined as:

$$L = \sum_{x=0}^{5} \sum_{i \in P_i} \theta_x \| D_i^{gt} - U_x(D^x)_i \|^2$$

$P_t$ represents the set of valid pixels in the ground truth depth map $D_{gt}$, $U_s$ is the bilinear interpolation operation that upsamples the predicted depth map at scale $x$, to the same resolution as the ground truth depth map $D_{gt}$ and $\theta_x$ is a hyperparameter used to balance the loss at each scale, which is empirically set to $4^{-x}$.

Under the six-scale architecture of the entire network, we perform normalization for each scale, which means that each scale is upsampled to the same resolution as the ground truth depth map through bilinear interpolation before calculating the error based on the L2 loss. Additionally, to fully supervise the training of the network, we set a weight that decreases with increasing resolution. This is because prediction errors at lower resolutions have a greater impact on the global structure, while prediction errors at higher resolutions are more reflected in local details.

**4. EXPERIMENTS**

1. Experimental Setup

We conducted comprehensive experiments on the KITTI depth completion dataset to evaluate the performance of our proposed method in outdoor scenes.

(1). Dataset

The KITTI Depth Completion dataset (DC) is collected by autonomous vehicles, with ground truth depth derived from temporally registered LiDAR scans and further validated through stereo image pairs. The entire KITTI Depth Completion dataset is divided into a training set of 86,898 frames, a validation set of 1,000 frames, and a test set of 1,000 frames. Since there are no valid LiDAR values near the top 100 pixels, we cropped the original data from 1216×352 to 1216×256 during training. For testing, we directly process the frames at the original resolution.

(2) Evaluation Metrics

We use four metrics to evaluate performance, including Root Mean Square Error (RMSE), Mean Absolute Error (MAE), Root Mean Square Error of the inverse depth (iRMSE), and Mean Absolute Error of the inverse depth (iMAE), where RMSE is selected as the primary evaluation metric by KITTI.

(3) Implementation Details

We implemented our method using PyTorch and trained it on devices equipped with 1 GPUs, each with $Nvidia\ RTX\ 4090$. We used $PointNet++$ to process the point cloud data obtained by back-projecting the sparse depth map, generating a 256-dimensional global feature vector. We then generated weights through a channel attention mechanism and combined these weights with the fused features output by U-Net to obtain the final fused features. Before using convolution to generate residual features to correct the initial dense depth map after obtaining the final fused features, we applied DroPath as a regularization technique. We used the AdamW optimizer with 0.05 weight decay. Additionally, we employed a Warem-up and cosine annealing strategy to adjust the learning rate, starting with an initial learning rate of 1/40, which is a fraction of the maximum learning rate. The learning rate gradually increases to the maximum learning rate over the first 10% training epochs and then begins cosine annealing over the remaining 90% training epochs, reducing the learning rate to 1/10 of the initial learning rate. We set the batch size to 2 and the maximum learning rate to 2.5e-4

2.Main Result

| Method | 2D | 3D | RMSE↓(mm) | MAE↓(mm) | iRMSE↓(1/km) | iMAE↓(1/km) | Publication |
|---|---|---|---|---|---|---|---|
| \multicolumn{8}{c}{KITTI} |
| CPSN | ✓ | | 1019.64 | 279.46 | 2.93 | 1.15 | ECCV 2018 |
| TWISE | ✓ | | 840.20 | 195.58 | 2.08 | 0.82 | CVPR 2021 |
| CSPN++ | ✓ | | 743.69 | 209.28 | 2.07 | 0.90 | AAAI 2020 |
| PENet | ✓ | | 730.08 | 210.5 | 2.17 | 0.94 | ICRA 2021 |
| RigNet | ✓ | | 712.66 | 203.25 | 2.08 | 0.90 | ECCV 2022 |
| LRRU | ✓ | | 696.51 | 189.96 | 1.87 | 0.81 | ICCV 2023 |
| BP-Net | ✓ | | **684.90** | 194.69 | 1.82 | 0.84 | CVPR 2024 |
| FuseNet | ✓ | ✓ | 752.88 | 221.19 | 2.34 | 1.14 | ICCV 2019 |
| ACMNet | ✓ | ✓ | 744.91 | 206.09 | 2.08 | 0.90 | TIP 2021 |
| PointFusion | ✓ | ✓ | 741.90 | 201.10 | 1.97 | 0.85 | ICCV 2021 |
| GraphCSPN | ✓ | ✓ | 738.41 | 199.31 | 1.96 | 0.84 | ECCV 2022 |
| PointDC | ✓ | ✓ | 736.07 | 201.87 | 1.97 | 0.87 | ICCV 2023 |
| DeCoTR | ✓ | ✓ | 717.07 | 195.30 | 1.92 | 0.84 | CVPR 2024 |
| TPVD | ✓ | ✓ | 693.97 | 188.60 | 1.82 | 0.81 | CVPR 2024 |
| GAC-Net | ✓ | ✓ | **680.82** | 193.85 | 1.81 | 0.84 | T-ITS 2025 |

Table 1. Quantitative results on the KITTI online depth completion leaderboard. 2D and 3D represent models utilizing 2D and 3D representations, respectively. Our GAC-Net achieves the best RMSE of 680.82 mm, significantly surpassing previous methods such as BP-Net (RMSE: 684.90 mm) and TPVD (RMSE: 693.97 mm). The best and second-best metrics are highlighted in the table.

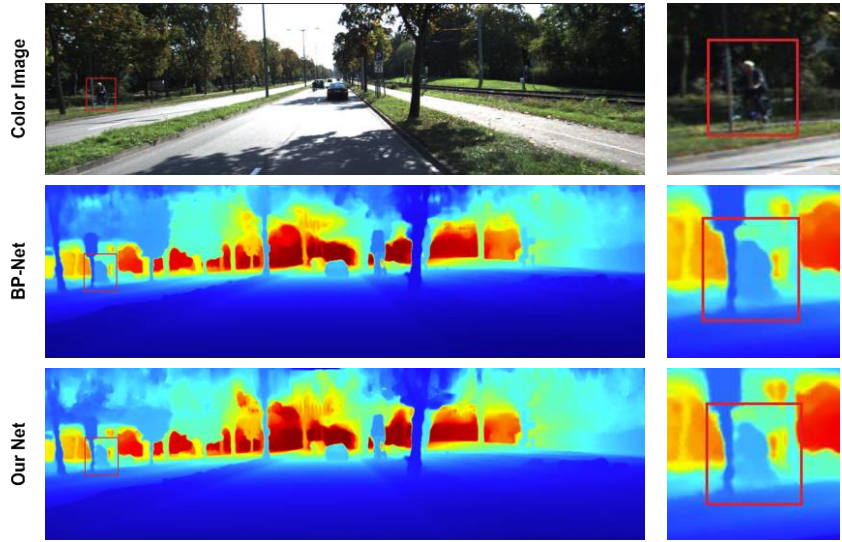

**Figure 3**, in the road scene on the left side of the image, in our network, the boundaries between the cyclist and the road sign are more distinctly separated, outperforming BPNet, demonstrating that our network achieves more refined and clear completion effects in complex boundaries.

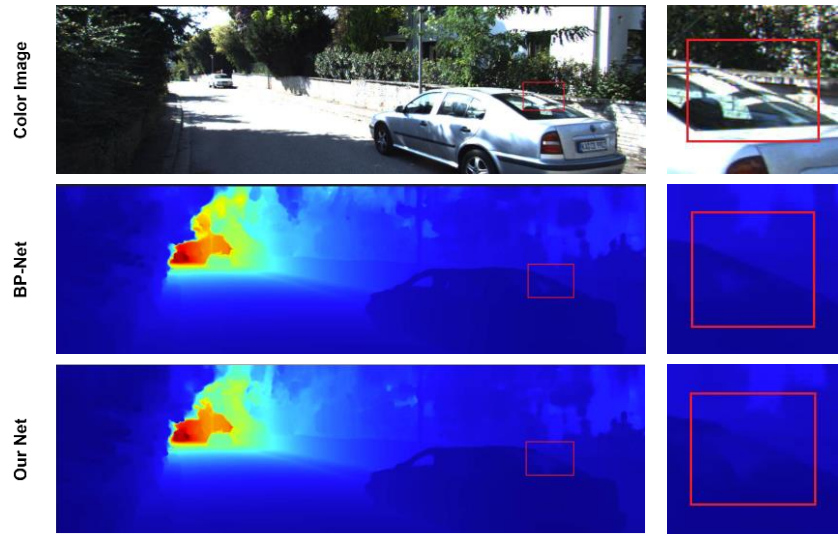

**Figure 4**, in scenarios where the smoothness of object surfaces and colors lead to sparse image textures and colors, as well as sparse original LiDAR data, our network's completion effect is superior to BPNet, indicating that our network has stronger robustness.

3. Ablation Studies

| GAC-Net | 2-stage | 3-stage | PointNet++ | concatenation | CAFFM | RMSE (mm) | MAE(mm) |
|---|---|---|---|---|---|---|---|
| i | ✓ | | | | | 730.08 | 210.55 |
| ii | | ✓ | | | | 714.4 | 205.63 |
| iii | | ✓ | ✓ | ✓ | | 697.5 | 198.64 |
| iv | | ✓ | ✓ | | ✓ | **680.82** | **193.85** |

Table 2. Ablation studies of our GAC-Net on KITTI validation split.

Table 2. presents the results of the ablation study on the KITTI validation set. The baseline model, GAC-Net-i, consists of only a two-stage network structure. After introducing the preprocessing stage, GAC-Net-ii reduces the RMSE from 730.08 mm to 719.6 mm. Based on GAC-Net-ii, we incorporated PointNet++ to extract global 3D features from point clouds and processed these features with initial fused features through feature concatenation, resulting in GAC-Net-iii. The inclusion of 3D features provided more comprehensive 3D geometric information, reducing the RMSE by 16.9 mm. Finally, replacing the feature concatenation in GAC-Net-iii with a channel attention-based feature fusion module further enhanced the network's representational capacity. As a result, the RMSE reached 680.82 mm, a reduction of 16.68 mm, achieving the current state-of-the-art (SOTA) performance.

**5. Conclusion**

In this paper, we propose a Geometry-Aware and Attention-Enhanced Multimodal Depth Completion Network (GAC-Net) to address the challenges of insufficient local geometric relationship modeling, missing global 3D structure, and limited effectiveness of 2D-3D feature fusion during the completion of sparse depth maps. Based on a three-stage network architecture, our network incorporates PointNet++ for global 3D feature extraction, enhancing its geometric modeling capabilities for the entire scene and effectively improving depth completion in large void areas and sparse scenes.

We further design a channel attention-based feature fusion module, which explicitly integrates global features to enhance multimodal fusion on top of the early fusion of image and depth features using U-Net. Finally, during the depth refinement stage, residual learning is employed to produce corrected depth maps, which, combined with the fine propagation of CSPN++, improves the accuracy of local detail and edge completion.

Experimental results demonstrate that the proposed method is effective in outdoor scenarios, such as on the KITTI dataset. Compared to state-of-the-art methods, our model generates more accurate and realistic depth estimations.